\documentclass[11pt]{article}

\usepackage[final]{acl}

\usepackage{times}
\usepackage{latexsym}

\usepackage[T1]{fontenc}

\usepackage[utf8]{inputenc}

\usepackage{microtype}

\usepackage{inconsolata}

\usepackage{graphicx}
\usepackage{amsmath}
\usepackage{tcolorbox}
\usepackage{amsfonts}
\usepackage{enumitem}
\usepackage{booktabs}
\usepackage{multirow}
\usepackage{makecell}
\usepackage{hyperref}
\usepackage{cleveref}
\usepackage[table]{xcolor}
\usepackage{xcolor}
\usepackage[version=4]{mhchem}
\title{ChemVLR: Prioritizing Reasoning in Perception for Chemical Vision-Language Understanding}

\author{
   Xuanle Zhao\textsuperscript{\rm 1,2}, Xinyuan Cai\textsuperscript{\rm 1}\thanks{Corresponding Authors.}, Xiang Cheng\textsuperscript{\rm 1}, Xiuyi Chen\textsuperscript{\rm 1}, Bo Xu\textsuperscript{\rm 1,2}\footnotemark[\value{footnote}] \\
 \textsuperscript{\rm 1} The Key Laboratory of Cognition and Decision Intelligence for Complex Systems,\\ Institute of Automation, Chinese Academy of Sciences \\
 \textsuperscript{\rm 2} School of Artificial Intelligence, University of Chinese Academy of Sciences \\
\texttt{zhaoxuanle2022@ia.ac.cn}
}

\begin{document}
\maketitle
\begin{abstract}
While Vision-Language Models (VLMs) have demonstrated significant potential in chemical visual understanding, current models are predominantly optimized for direct visual question-answering tasks. This paradigm often results in "black-box" systems that fail to utilize the inherent capability of Large Language Models (LLMs) to infer underlying reaction mechanisms. In this work, we introduce ChemVLR, a chemical VLM designed to prioritize reasoning within the perception process. Unlike conventional chemical VLMs, ChemVLR analyzes visual inputs in a fine-grained manner by explicitly identifying granular chemical descriptors, such as functional groups, prior to generating answers. This approach ensures the production of explicit and interpretable reasoning paths for complex visual chemical problems.
To facilitate this methodology, we implement a cross-modality reverse-engineering strategy, combined with a rigorous filtering pipeline, to curate a large-scale reasoning-and-captioning dataset comprising 760k high-quality samples across molecular and reaction tasks. Furthermore, we adopt a three-stage training framework that systemically builds model perception and reasoning capacity. Experiments demonstrate that ChemVLR achieves state-of-the-art (SOTA) performance, surpassing both leading proprietary models and domain-specific open-source baselines. We also provide comprehensive ablation studies to validate our training strategy and data generation designs. 
Code and model weights will be available at \url{https://github.com/xxlllz/ChemVLR}.

\end{abstract}

\section{Introduction}
Driven by the rapid progress in Reinforcement Learning with Verifiable Rewards (RLVR) \cite{guo2025deepseek,chen2025towards}, recent works have demonstrated robust reasoning capabilities across mathematical, programming, and scientific tasks \cite{zhang2025scientific,wang2025scireasoner}. These advancements position RLVR as a highly effective paradigm for enhancing structured reasoning in Large Language Models (LLMs). Motivated by this success, research in the chemical domain has begun shifting from purely Supervised Fine-Tuning (SFT) \cite{zhang2024chemllm,zhao2024chemdfm} to SFT-RL pipelines \cite{wang2025chem,zhao2025chemdfm}, aiming to incorporate explicit reasoning processes and improve performance on expert-level tasks. However, these approaches remain predominantly confined to the textual domain, thereby limiting their potential in addressing complex multimodal scenarios \cite{han2025generalist}.

In general vision-language domains, researchers have increasingly investigated RL-enhanced reasoning. Recent studies have pushed the boundaries of this field, demonstrating that RL significantly bolsters fine-grained visual understanding and complex cross-modal problem-solving \cite{shen2025vlm,huang2025vision}. However, in specialized domains such as multimodal chemistry, general-purpose VLMs often struggle to generalize due to a lack of domain-specific exposure. While specialized models like ChemVLM \cite{li2025chemvlm} and TinyChemVL \cite{zhao2025tinychemvl} have achieved competitive performance, they rely primarily on SFT within an end-to-end direct answering paradigm. These approaches do not fully leverage pretrained knowledge or elicit explainable reasoning processes, ultimately leading to suboptimal performance on complex tasks and limiting their utility as effective scientific research assistants.

In this work, we introduce ChemVLR, a pioneering framework that shifts the paradigm from holistic chemical perception to step-by-step visual reasoning. To enable this, we equip the model with a visual traversal mechanism, facilitating the precise analysis of complex chemical substructures before deducing the final answer. 
To overcome the bottleneck caused by the scarcity of high-quality reasoning data, we propose a Cross-Modality Reverse-Engineering strategy \cite{wang2025reverse}. By utilizing textual chemical queries paired with ground-truth answers, we employ advanced LLMs to reconstruct the underlying reasoning processes. Integrating retrieved IUPAC names \cite{long1983limit}, RDKit-computed functional groups \cite{bento2020rdkit}, and expert demonstrations as semantic anchors, the reasoning processes are generated abductively. Following rigorous filtering and verification strategies, including answer consistency checks and external LLM evaluation, as well as programmatic visualization \cite{zhao2025chartcoder}, we obtain a large-scale vision-based reasoning corpus. This dataset comprises 400k captions, 168k recognition samples, and 192k reaction prediction samples. Furthermore, our experiments find that open-source VLMs lack the general chemical image perception capacity. To address this challenge, our training methodology follows a progressive three-stage pipeline comprising Continual Pre-Training (CPT), SFT, and RL to cultivate domain expertise. Empirical results demonstrate that ChemVLR significantly outperforms existing baselines, especially the chemical-domain VLMs, achieving new SOTA performance. In summary, our contributions are as follows:
\begin{itemize}
    \item We propose a cross-modality reverse-engineering strategy to generate vision-based chemical reasoning data from textual queries. By integrating auxiliary semantic anchors and applying rigorous verification, we curate 760k high-quality samples across captioning, recognition, and prediction tasks.
    \item We introduce ChemVLR, the pioneering reasoning VLM for chemistry. It utilizes a progressive three-stage training pipeline to systematically cultivate the model's chemical perception and reasoning capabilities.
    \item We conduct extensive experiments and reveal that incorporating IUPAC nomenclature data can significantly activate pre-trained knowledge. Consequently, ChemVLR achieves SOTA performance, outperforming leading proprietary and chemical-domain VLMs.
\end{itemize}

\section{Related Works}
\subsection{Chemical Large Language Models} 
The application of LLMs to chemical challenges has emerged as a vibrant research frontier. Capitalizing on the generalization capabilities of foundation LLMs, pioneering works such as ChemLLM \cite{zhang2024chemllm} and ChemDFM \cite{zhao2024chemdfm} achieved success via SFT on curated chemical instruction datasets. However, the advent of reasoning-intensive models, exemplified by DeepSeek-R1 \cite{guo2025deepseek}, has catalyzed a paradigm shift toward enhancing complex problem-solving capabilities. Consequently, recent studies \cite{narayanan2025training, li2025mol, zhao2025molreasoner} have pivoted toward leveraging Chain-of-Thought (CoT) data and RL to elicit long-term interpretable reasoning processes. For example, models like ChemDFM-R \cite{zhao2025chemdfm} and Chem-R \cite{wang2025chem} enhance the reasoning capacity of LLM by integrating three-phase training, including domain pre-training, reasoning-oriented SFT, and RL.

\begin{figure*}
    \centering
    \includegraphics[width=0.98\linewidth]{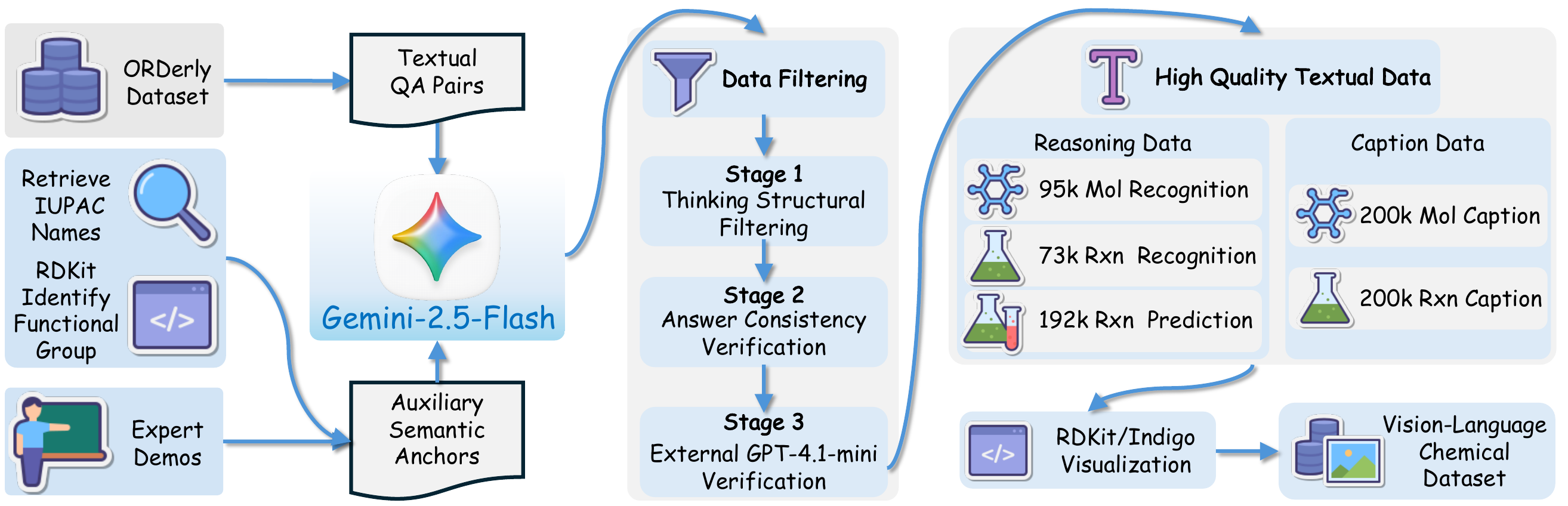}
    \vspace{-5pt}
    \caption{Overview of our proposed Cross-Modality Reverse-Engineering strategy for data generation. The pipeline begins with an initial dataset constructed from textual SMILES QA pairs, enriched with auxiliary information, including IUPAC names, functional groups, and expert demonstrations. Once Gemini-2.5-Flash generates the reasoning and caption data, they are subjected to a rigorous three-stage filtering and verification process to ensure accuracy and quality. Subsequently, the filtered data are rendered into images using RDKit and Indigo code.}
    \label{fig:data_generation}
    \vspace{-12pt}
\end{figure*}

\subsection{Vision Language Models}
While generalist models like GPT-4o \cite{openai2024gpt4o} have established strong baselines for visual tasks, the unique complexity of molecular structures has necessitated domain-specific adaptations. ChemVLM \cite{li2025chemvlm} addresses this by constructing specialized visual-chemical datasets and utilizing the chemical LLM backbone for enhanced domain adaptation. ChemDFM-X \cite{zhao2024chemdfmx} introduces distinct encoders to achieve holistic alignment between visual, graph, and textual representations. In the generative domain, ChemMLLM \cite{zhang2024chemllm} employs VQ-GAN \cite{esser2021taming} to facilitate both molecular image understanding and generation. Recently, TinyChemVL \cite{zhao2025tinychemvl} optimizes computational efficiency through adaptive token merging and pruning and incorporates vision-based reaction-level tasks to broaden the task scope. However, these existing approaches primarily rely on direct SFT, leaving the explicit, explainable reasoning capacity of the models largely unexplored.
While in the general vision-language domain, recent works focus on enhancing the long-chain reasoning capabilities of VLMs. By adapting Group Relative Policy Optimization (GRPO) \cite{shao2024deepseekmath} to multimodal scenarios, VLMs such as Vision-R1 \cite{huang2025vision} and R1-OneVision \cite{yang2025r1} have successfully unlocked long-term reasoning abilities in visual tasks.

\section{Dataset Construction}
We introduce a cross-modality reverse-engineering strategy to construct the training dataset, integrating three core data types: reasoning, captions, and instructions.

\subsection{Reasoning Data Generation}
Despite the abundance of public corpora for molecular structures and reactions, there is a marked scarcity of datasets annotated with explicit reasoning traces, particularly for visually grounded tasks. Consequently, existing VLMs exhibit limited capacities in fine-grained visual molecular understanding and reaction prediction.
To address this, inspired by recent advances \cite{wang2025reverse}, we propose a cross-modality reverse-engineering strategy to synthesize reasoning process from textual queries and ground-truth answers. Specifically, we categorize vision-based chemical reasoning tasks into recognition (molecular/reaction) and prediction (reaction), where all targets are standardized as SMILES strings.
Leveraging textual corpora from the ORDerly dataset \cite{wigh2024orderly}, we construct reasoning-oriented instructions and employ Gemini-2.5-Flash \cite{comanici2025gemini} to deduce reasoning processes by simulating visual perception patterns. For recognition tasks, the queries and answers are both molecular and reaction SMILES. For prediction tasks, queries consist of reactants, reagents, and solvents, while answers correspond to the products. All chemical entities are formatted as SMILES strings.

\begin{table}[t]
\centering
\small
\setlength{\tabcolsep}{6pt}
\begin{tabular}{lccc}
\toprule
\textbf{Setting} & \textbf{Mol. Rec.} & \textbf{Rxn. Rec.} & \textbf{Rxn. Pred.} \\
\midrule
SMILES          & 78.0\% & 58.0\% & 55.0\% \\
~~+ Demo               & 80.0\% & 64.0\% & 62.0\%   \\
~~~~+ IUPAC     & 92.0\% & 71.0\% & 71.0\% \\
~~~~~~+  RDKit & \textbf{95.0\%} & \textbf{73.0\%} & \textbf{76.8\%} \\
\bottomrule
\end{tabular}
\vspace{-5pt}
\caption{Effectiveness comparison of various data generation settings. We generate 10k samples for each category utilizing our reverse-engineering strategy, except for the last row which utilizes the whole reasoning set. The percentages represent the retention rate of high-quality data following the three-stage filtering protocol.}
\label{tab:comparison}
\vspace{-15pt}
\end{table}

\begin{figure*}
    \centering
    \includegraphics[width=0.98\linewidth]{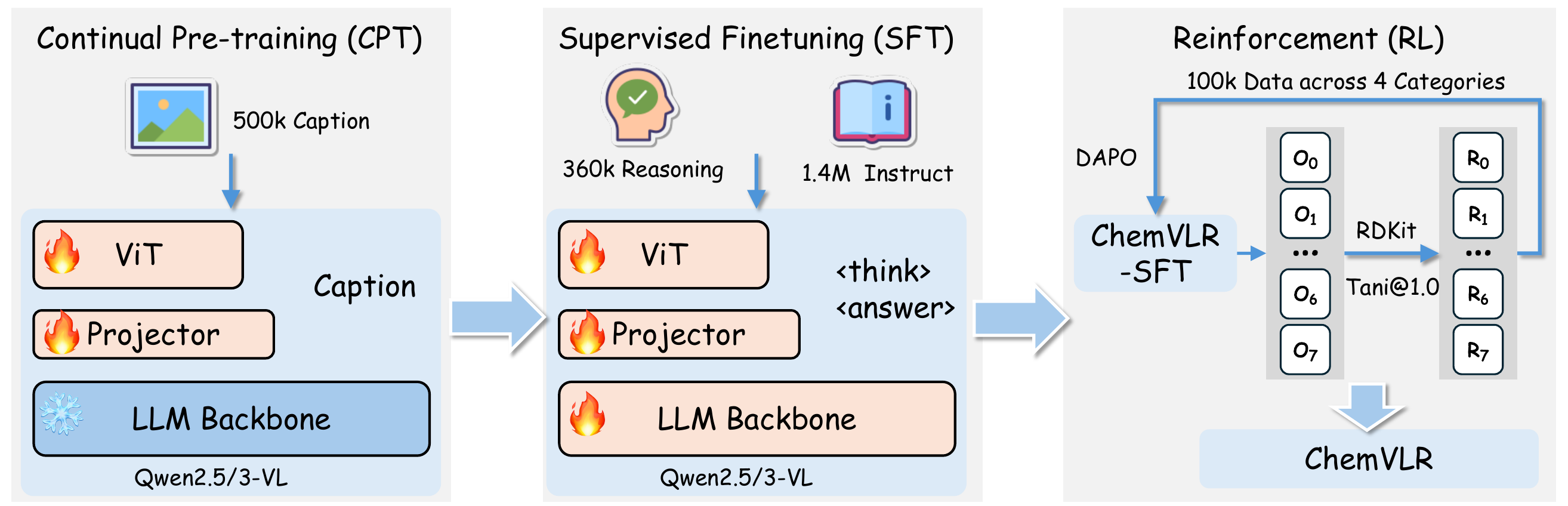}
    \vspace{-5pt}
    \caption{The training framework of ChemVLR proceeds in three stages. First, we conduct chemical-domain continual pre-training using caption data. Second, during the SFT stage, we train on a mixture of reasoning and instruction data. For the RL stage, we retain only non-trivial yet solvable instances and optimize utilizing DAPO.}
    \label{fig:chemvlr}
    \vspace{-13pt}
\end{figure*}

Our preliminary experiments reveal that providing only SMILES sequences is insufficient for Gemini-2.5-Flash to synthesize accurate reasoning processes. To address this, we propose integrating auxiliary semantic anchors to incorporate domain expertise, comprising retrieved IUPAC names, RDKit-computed functional groups, and expert demonstrations. Specifically, we retrieve corresponding IUPAC names from PubChem \cite{kim2023pubchem} and patent databases, and utilize RDKit to identify functional groups. Furthermore, we manually curate expert demonstrations to exemplify step-by-step visual analysis, systematically identifying functional groups and potential reaction sites.
Empirical results demonstrate that this augmented approach significantly enhances both the efficiency and accuracy of the generated reasoning processes (as shown in Table \ref{tab:comparison}). Leveraging this strategy, we construct a large-scale dataset of 450k vision-based reasoning samples, comprising molecular recognition (100k), reaction recognition (100k), and reaction prediction (250k).

To guarantee the high fidelity of the synthesized dataset, we implement a rigorous three-stage filtering protocol. First, we perform structural filtering, retaining only traces that exhibit explicit visually grounded reasoning patterns. Second, we enforce answer consistency by validating the final answer derived from Gemini-2.5-Flash against the ground-truth SMILES, eliminating any discrepancies. Finally, we conduct external reasoning verification using GPT-4.1-mini as an independent verifier.By providing both the task query and synthesized reasoning processes, we retain only those samples where the rationale successfully guides the verifier to recover the correct ground truth. This process yields 360k high-quality samples, distributed across molecular recognition (95k), reaction recognition (73k), and reaction prediction (192k). We provide the thinking length statistics of in Table~\ref{tab:reasoning_stats}. 

\subsection{Caption Data Generation}
Our empirical analysis reveals that general open-source VLMs, such as Qwen3-VL \cite{bai2025qwen3vltechnicalreport} and InternVL3.5 \cite{wang2025internvl3}, exhibit significant limitations in domain-specific chemical visual understanding, particularly in discerning fine-grained substructures like functional groups. To address this deficiency, we extend our reverse-engineering strategy to curate a fine-grained image captioning dataset tailored for continual pre-training (CPT). Leveraging a distinct dataset complementary to our reasoning corpus, we generate captions conditioned on IUPAC names and RDKit-derived functional groups. We ensure data quality through the a reconstruction-based filtering protocol after structural filtering, employing GPT-4.1-mini to retain only those captions from which the ground-truth SMILES can be accurately recovered. This process yields a curated dataset comprising 200k molecular and 200k reaction image-caption pairs.

\subsection{Instruct Data Generation}
Our preliminary SFT experiments relying solely on reasoning data revealed that model performance remains suboptimal on relevant visual chemical tasks. To mitigate this limitation, we adopt the data construction strategy from TinyChemVL \cite{zhao2025tinychemvl}, reformulating samples from the ORDerly dataset into a direct instruction-following format. This yields a comprehensive set sourced from existing datasets \cite{morin2023molgrapher,qian2023molscribe,tan2025chemmllm}, comprising 470k samples for molecular recognition, 200k for reaction recognition, and 400k for reaction prediction. Additionally, we introduce an image-to-IUPAC conversion task. During the generation of reasoning and captioning data, we observe that incorporating IUPAC names significantly enhances the model's comprehension of molecular structures and reaction types. We attribute this improvement to the fact that IUPAC nomenclature appears more frequently in general pre-training corpora compared to other chemical representations. To this end, we construct an additional 300k image-to-IUPAC samples, expanding our direct instruction corpus to 1.4M samples.

\begin{table}[t]
  \centering
  \small
  \begin{tabular}{lccc}
    \toprule
    \textbf{Caption Dataset} & \textbf{Samples} & \textbf{Average} & \textbf{SD} \\
    \midrule
    ChEBI-MM & 26k & 151.57 & 80.46 \\
    ChemMLLM & 70k & 95.92 & 52.46 \\
    Molecule (Ours) & 200k & 409.93 & 139.03 \\
    Reaction (Ours) & 200k & 441.06 & 76.85 \\
    \bottomrule
  \end{tabular} 
  \vspace{-5pt}
  \caption{Token Length Statistics of Caption Datasets. The mean and standard deviation (SD) of token counts are computed via the Qwen3-VL tokenizer.}
  \label{tab:caption_stats}
  \vspace{-15pt}
\end{table}

\section{ChemVLR Model}
With the dataset constructed, we leverage these samples to enhance the visual chemical understanding capabilities of VLMs through a three-stage training pipeline comprising CPT, SFT, and RL.

\subsection{Continual Pre-training}
Our preliminary experiments reveal that current open-source VLMs \cite{wang2025internvl3, bai2025qwen3vltechnicalreport} exhibit limited generalization capabilities in the chemical domain, often failing to accurately parse molecular structures and reaction processes. To bridge this modality gap, we introduce a chemical-specific alignment stage prior to SFT. Following established pre-training paradigms \cite{coreteam2025mimovltechnicalreport, dong2025scalable}, we perform Continual Pre-training (CPT) by optimizing the visual encoder (ViT) and projector layer while keeping the LLM backbone frozen. This stage utilizes a compiled dataset of approximately 500k chemical image-text pairs, comprising our constructed 400k molecular and reaction captions, supplemented by 26k samples from ChEBI-20-MM \cite{liu2025quantitative} and 70k from ChemMLLM \cite{tan2025chemmllm}. This approach effectively adapts visual perception to chemical structures while preserving the model's pre-trained linguistic capabilities.

\subsection{Large-scale Supervised Fine-tuning}
Subsequent to CPT, we perform SFT on our constructed large-scale dataset to enhance the model's reasoning and chemical instruction-following capacity. We adopt a structured formatting protocol wherein reasoning traces and answers are encapsulated within \texttt{<think>} and \texttt{<answer>} tags. Similarly, final outputs are explicitly demarcated using \texttt{<SMILES>} and \texttt{<IUPAC>} delimiters. The model is trained via the standard autoregressive objective.
\begin{equation}
\small{
\mathcal{L}(\theta):=-\mathbb{E}_{(q, a) \sim \mathcal{D}_\text{SFT}} \sum_{t=1}^T \log P\left(a_t \mid q, a_{<t} ; \theta\right),
}
\end{equation}
where $(q, a)$ denotes a query-answer pairs from dataset $\mathcal{D}_\text{SFT}$, comprising 360k reasoning and 1.4M instruction samples.

\subsection{Reinforcement Learning}
Despite SFT establishing a strong baseline, models trained under this paradigm remain vulnerable to noise and hallucinations inherent in the training data, often exhibiting brittle generalization in complex reasoning tasks. To transcend these limitations and further refine the policy, we adopt optimize the model using Decoupled Clip and Dynamic Sampling Policy Optimization (DAPO) \cite{yu2025dapo}. Diverging from GRPO \cite{shao2024deepseekmath}, DAPO explicitly integrates token-level loss computation and an asymmetric decoupled clipping strategy to enhance training stability.

\paragraph{DAPO.}
For each input $(q, a)$, the policy $\pi_\theta$ samples a group of $G$ candidate responses $\{o_i\}_{i=1}^G$. Each response receives a reward $R_i$ as described in the next paragraph. We optimize the policy objective with a decoupled, asymmetric clipping mechanism. The  objective function is formulated as:
\begin{equation}
\small{
\begin{split}
\mathcal{J}_{\text{DAPO}}(\theta) = \mathbb{E}_{(q,a)\sim\mathcal{D}_\text{RL}, \{o_i\}_{i=1}^G\sim\pi_{\theta_{\text{old}}}(\cdot|q)} \qquad\qquad \\
\left[ \frac{1}{\sum_{i=1}^G |o_i|} \sum_{i=1}^G \sum_{t=1}^{|o_i|} \min \left( r_{i,t}(\theta)\hat{A}_{i,t}, \right. \right. \\
\left. \left. \text{clip}\left(r_{i,t}(\theta), 1 - \varepsilon_{\text{low}}, 1 + \varepsilon_{\text{high}}\right)\hat{A}_{i,t} \right) \right]
\end{split}
}
\end{equation}
where the probability ratio $r_{i,t}(\theta)$ and the group-normalized advantage $\hat{A}_{i}$ are defined as,
\begin{equation}
\small{
\hat{A}_{i,t} = \frac{R_i - \text{mean}(\{R_j\}_{j=1}^G)}{\text{std}(\{R_j\}_{j=1}^G)},
}
\end{equation}
\begin{equation}
\small{
r_{i,t}(\theta) = \frac{\pi_\theta(o_{i,t} \mid q, o_{i,<t})}{\pi_{\theta_{\text{old}}}(o_{i,t} \mid q, o_{i,<t})}.
}
\end{equation}

\begin{table*}[t]
\setlength{\tabcolsep}{4pt} 
\centering
\resizebox{\textwidth}{!}{
\begin{tabular}{lcccccccc}
\toprule
\multirow{2}{*}{\textbf{Models}} & \multirow{2}{*}{\textbf{Paradigm}} & \multicolumn{2}{c}{\textbf{MMChemOCR}} & \multicolumn{2}{c}{\textbf{img2smiles}} & \multicolumn{1}{c}{\textbf{ChemRxn-V$_\text{R}$}} & \multicolumn{2}{c}{\textbf{ChemRxn-V$_\text{P}$}} \\
\cmidrule(r){3-4} \cmidrule(r){5-6} \cmidrule(r){7-7} \cmidrule(l){8-9}
& & Avg Sim. & Tani\text{@}1.0 & Avg Sim. & Tani\text{@}1.0 & Tani\text{@}1.0 & Avg Sim. & Tani\text{@}1.0 \\
\midrule
\multicolumn{9}{l}{\textbf{Proprietary Models}} \\ zz
Gemini-3-Flash & - & 77.6 & 61.2 & 74.5 & 63.8 & 4.4 & 76.8 & 51.7 \\
GPT-5-mini  & - & 28.8 &  5.7 & 20.4 & 5.1 & 0.7 & 24.3 & 2.3 \\
GPT-4o & - & 36.8& 3.4& 29.0& 0.1 & 0.1& 30.4& 1.4\\
\midrule
\multicolumn{9}{l}{\textbf{Open-Source General-Domain Models}} \\
Phi-3.5-Vision & Instruct & 0.4& 0.0& 1.2& 0.0 & 0.0& 0.8& 0.0\\
Qwen2.5-VL-7B & Instruct & 25.5& 0.4& 28.2& 0.0 & 0.1& 11.9& 0.0\\
InternVL3.5-8B & Instruct & 85.4 & 59.1 &  15.1 & 4.4 & 1.9 &39.4 & 5.9\\
Qwen3-VL-8B  & Instruct & 26.3 & 10.2 &  27.9 & 1.3 & 0.3 & 18.2 & 0.1\\
InternVL3.5-14B & Instruct w/ CoT & 86.1 & 69.0 &  17.8 & 6.8 &0.9 &28.7 & 2.8\\
InternVL3.5-38B & Instruct w/ CoT & 78.0 & 40.1 &  29.1& 9.0 & 2.1 &37.0 & 4.9\\
\midrule
\multicolumn{9}{l}{\textbf{Chemical-Domain VLMs}} \\ 
ChemVLM-8B & Instruct & 81.7& 57.7& 55.0& 15.0& 0.0& 4.8& 0.0\\
ChemDFM-X  & Instruct & 70.9& 36.5& 90.9& 77.6& 3.2& 12.7& 0.7\\
TinyChemVL & Instruct & 91.2& 77.4& 89.5& 75.6& 67.9& 78.9& 52.4\\
\midrule
\rowcolor[HTML]{D7F6FF} ChemVLR-7B & Thinking & 93.2 & 83.9 & \textbf{97.8} & \textbf{92.8} & \textbf{74.6} & \textbf{84.9} & 67.2 \\
\rowcolor[HTML]{D7F6FF} ChemVLR-8B & Thinking & \textbf{93.8} & \textbf{84.6} & 97.4 & 92.7 & 74.4 & 84.8 & \textbf{67.8} \\
\bottomrule
\end{tabular}
}
\vspace{-5pt}
\caption{Combined evaluation on molecular and reaction tasks.
Avg Sim. and Tani@1.0 denote Average Tanimoto Similarity and Tanimoto Hit@1.0. 
ChemRxn-V$_\text{R}$ and ChemRxn-V$_\text{P}$ correspond to the reaction recognition and prediction subsets. 
\textbf{Bold} indicates the best performance.
}
\vspace{-15pt}
\label{tab:main_results}
\end{table*}

\paragraph{Reward Design.} 
To facilitate effective RL training, we design verifiable rewards tailored for the vision-based chemical reasoning task. We adopt a widely used composite reward formulation that comprises accuracy and format components. Specifically, the accuracy reward validates the correctness of the final output, while the format reward enforces the structural adherence of the reasoning process.
\begin{itemize}[leftmargin=*]
    \item \textbf{Accuracy Reward.} We utilize different reward functions for SMILES and IUPAC output tasks. For SMILES, we use fingerprint-based Tanimoto similarity to handle canonicalization variants, granting a reward of 1 only for a 1.0 similarity score. For IUPAC outputs, we adopt exact string matching to mitigate the reliance on potentially unstable external PubChem services.
    \item \textbf{Format Reward.} We employ a regex-based mechanism to enforce structural compliance, validating that generated sequences are strictly encapsulated within \texttt{<think>} and \texttt{<answer>} tags.
\end{itemize}
We apply binary scoring to all reward components, calculating the overall reward as a weighted average of accuracy and formatting metrics. In this stage, we optimize the model across four visual tasks: molecular recognition, reaction recognition, reaction prediction, and molecule-to-IUPAC translation. However, identical rewards across rollouts cause relative advantages to vanish, rendering optimization ineffective. To address this, we employ the SFT model to filter for instances of moderate difficulty. Specifically, we source unseen samples from the ORDerly dataset and generate 4 independent rollouts for each. We retain only instances that yield divergent outcomes, discarding both trivial and impossible cases. Ultimately, this strategy yields a curated dataset of 100k samples, comprising approximately 25k instances for each task.

\section{Experiments}
\subsection{Implementation Details}
For the data generation process, we utilize Gemini-2.5-Flash \cite{comanici2025gemini} to reverse-engineer the reasoning process and employ GPT-4.1-mini \cite{openai2025gpt41} for data verification. For the training setup, we choose Qwen2.5-VL-7B \cite{bai2025qwen2} and Qwen3-VL-8B-Instruct \cite{bai2025qwen3vltechnicalreport} as our backbones. During the CPT stage, we optimize the ViT and projector layer with a global batch size of 64. Subsequently, for the SFT and RL stages, we perform full-parameter fine-tuning with global batch sizes of 64 and 128, respectively. All stages are executed for a single epoch on a cluster of 16 NVIDIA H800 GPUs.

\subsection{Baselines and Benchmarks}
To compare the capacity of our ChemVLR, we choose baseline models from three setups: (1) Proprietary models, including GPT-5-mini \cite{openai2025gpt5}, Gemini-3-Flash \cite{googledeepmind2025gemini3flash} and GPT-4o \cite{openai2024gpt4o}. (2) Open-Source VLMs, including Phi-3.5-Vision \cite{abdin2024phi}, Qwen2.5-VL (7B) \cite{bai2025qwen2}, InternVL3.5 (8B, 14B, 38B) \cite{wang2025internvl3}, and Qwen3-VL (8B-Thinking/Instruct) \cite{bai2025qwen3vltechnicalreport}. (3) Chemical-Domain VLMs, including ChemVLM-8B \cite{li2025chemvlm}, ChemDFM-X \cite{zhao2024chemdfmx}, TinyChemVL \cite{zhao2025tinychemvl}. 
All baselines are evaluated on MMChemOCR \cite{li2025chemvlm} and img2smiles \cite{tan2025chemmllm} for molecular recognition, as well as ChemRxn-V \cite{zhao2025tinychemvl} for reaction recognition and prediction.

\begin{figure*}
    \centering
    \includegraphics[width=0.98\linewidth]{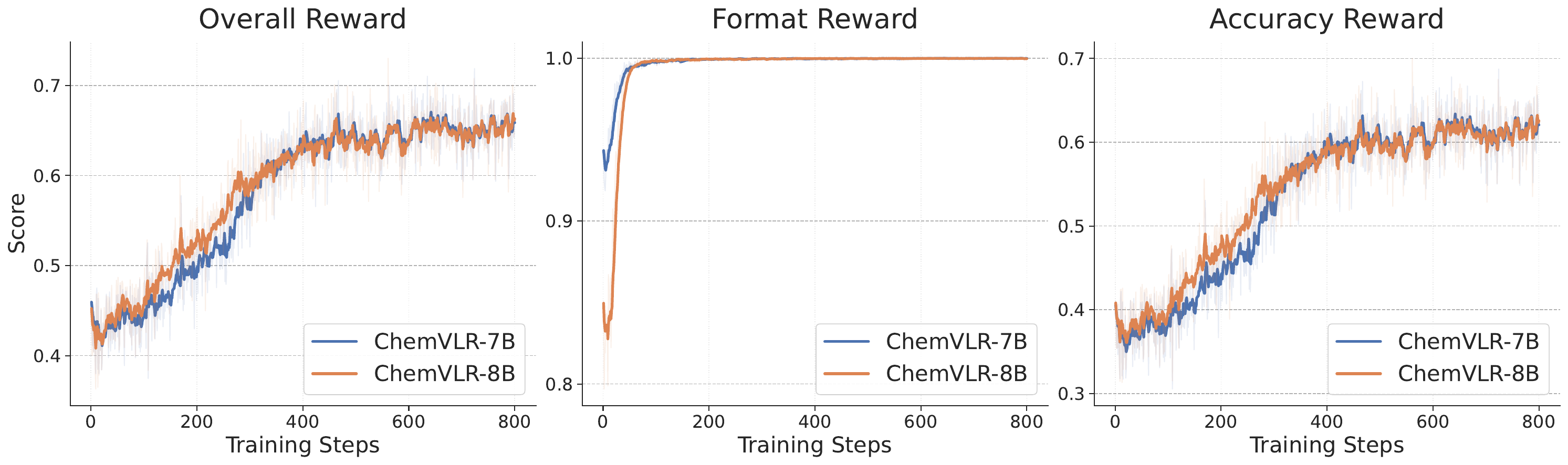}
    \vspace{-5pt}
    \caption{The reward curve during the RL training process. The overall rewards are the weighted average of format and accuracy rewards.}
    \label{fig:rl_curve}
    \vspace{-10pt}
\end{figure*}

\subsection{Main Results}
As presented in Table \ref{tab:main_results}, ChemVLR achieves SOTA performance across all benchmarks, significantly outperforming the previous leading models. We also evaluate recently released proprietary and open-source models, including Gemini-3-Flash and the Qwen3-VL series. While these generalist models demonstrate improved understanding and reasoning over prior baselines, they significantly underperform VLMs specifically adapted to the chemical domain. This gap highlights that, unlike general-domain queries, chemical tasks require specialized domain expertise for effective reasoning. Notably, on the MMChemOCR benchmark, ChemVLR is the first VLM to achieve parity with specialized SMILES OCR models such as Decimer \cite{rajan2021decimer} (Avg. Sim. 85.0, Tani@1.0 77.3) and Molscribe \cite{qian2023molscribe} (Avg. Sim. 92.0, Tani@1.0 89.1), while simultaneously providing detailed functional group analysis. This demonstrates that versatile VLMs can match the precision of specialist models without compromising broader capabilities. Furthermore, the substantial improvement over previous instruction-tuned chemical VLMs confirms that integrating reasoning enhances recognition. 

Figure \ref{fig:rl_curve} illustrates the training dynamics of the RL process, validating the effectiveness of this stage. Notably, we observe a marked surge in rewards between steps 200 and 400. This phenomenon is particularly pronounced in ChemVLR-7B, which exhibits a distinct ''Aha Moment'', signaling emergent reasoning capabilities.

\subsection{Ablation Study}
We conduct comprehensive ablation studies to validate the efficacy of our proposed model and data strategies. This analysis addresses the following research questions: (1) \textit{What is the specific contribution of each training stage to the overall performance?} (2) \textit{How do reasoning and instruction data distinctively impact the SFT process?} (3) \textit{How do different reward function formulations influence the RL stage?}

\paragraph{Training Stages.} 
Table \ref{tab:ablation_stage} summarizes the results of varying training stage combinations. We conduct comprehensive single and multi-stage ablation studies using Qwen3-VL-8B-Instruct as the baseline. The results demonstrate that Chem-VLR-8B achieves optimal performance when incorporating all training stages. Specifically, integrating CPT prior to SFT enhances visual chemical comprehension compared to the SFT-only baseline, while augmenting the CPT-SFT model with RL further elevates reasoning capabilities, yielding an average improvement of 9\% across all tasks. Notably, applying RL in isolation yields negligible gains, as the generalist baseline lacks the foundational chemical domain understanding required for effective reinforcement learning.

\begin{figure*}
    \centering
    \includegraphics[width=0.98\linewidth]{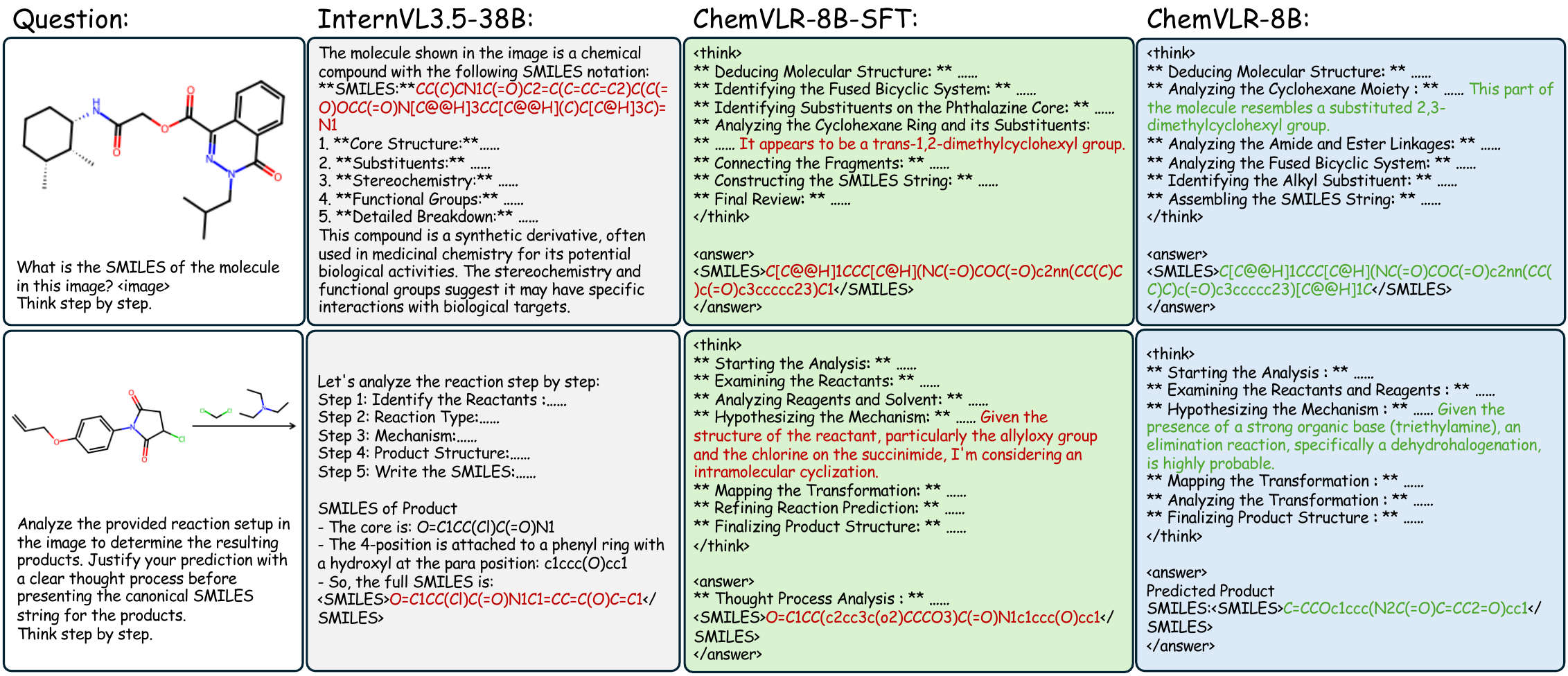}
    \vspace{-5pt}
    \caption{Showcase of the reasoning process and final answer on the molcular recognition and reaction prediction tasks. The ChemVLR-8B generate the correct answer compare with other baselines.}
    \label{fig:showcase}
    \vspace{-10pt}
\end{figure*}

\begin{table}[t]
\centering
\small
\setlength{\tabcolsep}{5pt}
\begin{tabular}{lccc}
\toprule
\textbf{Ablation} & \textbf{img2smiles} & \textbf{Rxn-V}$_\textbf{R}$ & \textbf{Rxn-V}$_\textbf{P}$ \\
\midrule
SFT only & 81.5 & 60.2 & 56.8 \\
RL only & 2.1 & 0.2 & 0.1 \\ 
\midrule
CPT + SFT & 84.7 & 63.5 &  59.3 \\
ChemVLR-8B & 92.7 & 74.4 & 67.8 \\
\bottomrule
\end{tabular}
\vspace{-5pt}
\caption{Ablation study of different training stages. Experiments are conducted using Qwen3-VL-8B-Instruct, with performance evaluated via Tani@1.0. For brevity, we refer to ChemRxn-V as Rxn-V.}
\label{tab:ablation_stage}
\vspace{-10pt}
\end{table}

\begin{table}[t]
\centering
\small
\setlength{\tabcolsep}{5pt}
\begin{tabular}{lccc}
\toprule
\textbf{Ablation} & \textbf{img2smiles} & \textbf{Rxn-V}$_\textbf{R}$ & \textbf{Rxn-V}$_\textbf{P}$ \\
\midrule
Full SFT Data & 84.7 & 63.5 & 62.6 \\
w/o IUPAC & 80.8 & 56.2 & 56.1\\
w/o Reasoning & 76.4 & 53.8 & 53.3 \\
w/o Instuction & 68.4 & 46.5 & 46.9 \\
\bottomrule
\end{tabular}
\vspace{-5pt}
\caption{Ablation study on SFT data composition. The instruction data contains the molecular-to-IUPAC data.}
\label{tab:ablation_data}
\vspace{-10pt}
\end{table}

\paragraph{SFT Data Composition.} Our experiments demonstrate that integrating diverse data types significantly enhances SFT performance, with molecule-to-IUPAC data emerging as a critical component. To investigate this, we conduct ablation studies on the CPT model, specifically evaluating the impact of excluding molecule-to-IUPAC, reasoning, and instruction data. The results in Table \ref{tab:ablation_data} reveal that removing vision-based Molecule-to-IUPAC data causes a substantial performance decline. This suggests that incorporating IUPAC-related data is essential to surmount the performance plateau inherent in training exclusively on SMILES tasks. We attribute this to the capacity of IUPAC nomenclature to bridge the modality gap, aligning visual features with the rich chemical knowledge encoded in the LLM during pre-training, thereby significantly augmenting reasoning capabilities. Furthermore, the exclusion of either instruction or reasoning data also leads to marked performance drops, validating the efficacy of our mixed data strategy.

\paragraph{RL Reward.} Complementing our ablations on training stages and SFT data, we investigate the impact of the RL reward formulation. We evaluate our proposed structural identity reward, which triggers solely when Tanimoto similarity equals 1.0, against two alternatives: the continuous Tanimoto similarity score and naive exact string matching (ignoring structural equivalence). As reported in Table \ref{tab:ablation_reward}, the structural identity reward outperforms both variants, yielding the optimal performance.

\begin{table}[t]
\centering
\small
\setlength{\tabcolsep}{5pt}
\begin{tabular}{lccc}
\toprule
\textbf{Ablation} & \textbf{img2smiles} & \textbf{Rxn-V}$_\textbf{R}$ & \textbf{Rxn-V}$_\textbf{P}$ \\
\midrule
Tani Simarity & 90.2 & 70.8 & 65.9 \\
Exact Matching & 90.8 & 69.9 & 66.4 \\
Struct-ID (Ours) & 92.7 & 74.4 & 67.8 \\
\bottomrule
\end{tabular}
\vspace{-5pt}
\caption{Ablation study on different reward functions for RL. We compare our utilized structural identity reward against dense Tanimoto similarity and sparse exact string matching rewards.}
\label{tab:ablation_reward}
\vspace{-10pt}
\end{table}

\subsection{Case Study Analysis}
To evaluate reasoning capabilities, Figure \ref{fig:showcase} visualizes a case study of molecular recognition and reaction prediction, comparing the reasoning process of InternVL3.5-38B, ChemVLR-8B-SFT, and ChemVLR-8B. 
In the reaction prediction task, ChemVLR-8B accurately identifies the mechanism as a base-mediated dehydrohalogenation. It correctly deduces that triethylamine (\ce{Et_3N}) abstracts the acidic $\alpha$-proton from the chlorosuccinimide to facilitate \ce{HCl} elimination, yielding the maleimide core. Notably, it also demonstrates chemoselectivity by explicitly stating that the allyloxy ether moiety remains intact under these mild conditions.
In contrast, ChemVLR-8B-SFT exhibits a specific reasoning failure characterized by over-interpretation. While it correctly identifies the initial elimination step (i.e., the loss of \ce{HCl}), it erroneously hypothesizes a subsequent complex intramolecular cyclization. The model incorrectly predicts that the reactive core interacts with the allyloxy tether (\ce{-OCH_2CH=CH_2}) to yield a dihydrobenzofuran derivative. It fails to recognize that the reaction thermodynamically terminates at the maleimide stage, erroneously proposing a coupling between the alkene and the ether moiety.
InternVL3.5-38B exhibits a fundamental perception and reasoning failure. It mischaracterizes the solvent, dichloromethane (\ce{CH_2Cl_2}), as a reactive species and erroneously postulates the cleavage of the chemically inert allyloxy ether moiety. Conseqconstructs a reaction pathway that completely overlooks the thermodynamically ffavoured dynamically favored elimination mechanism.
The complete model responses with the thinking processes are denoted in the \Cref{fig:mol_rec_internvl3_5_38b,fig:mol_rec_chemvlr_8b_sft,fig:mol_rec_chemvlr_8b,fig:rxn_pred_internvl3_5_38b,fig:rxn_pred_chemvlr_8b_sft_1,fig:rxn_pred_chemvlr_8b_sft_2,fig:rxn_pred_chemvlr_8b}.  

\section{Conclusion}
In this work, we propose ChemVLR, the first reasoning-capable chemical VLM to achieve superior performance on vision-based chemical tasks. To train this model, we introduce a cross-modality reverse-engineering strategy. By adhering to a rigorous filtering pipeline, we construct a high-quality dataset comprising both reasoning chains and image captions. We believe our proposed framework establishes a robust baseline for upcoming studies in multimodal chemical tasks.

\section*{Limitation}
In this work, we propose a large-scale reasoning dataset to enhance the model's thinking capacity. Despite our rigorous three-stage filtering strategy, a marginal proportion of the data still contains reasoning errors. We currently lack a more effective mechanism for further data refinement.
Currently, our focus remains on recognition and reaction prediction. For other tasks such as property prediction, constructing coherent reasoning processes is non-trivial. Determining how to generate these reasoning chains remains an under-explored area, we hypothesize that leveraging code-centric approaches could be a promising strategy. Another limitation is the lack of consideration for real-world visual scenarios, such as those found in K-12 educational tasks. Due to current data source constraints, we intend to address these specific tasks in our future research.

\section*{Ethic Statement}
In this work, we utilize Gemini to generate reasoning processes for chemical reactions. Despite rigorous filtering, residual erroneous data may persist, potentially leading to incorrect model outputs. To mitigate these risks, we emphasize that manual verification of reasoning correctness is essential when utilizing the dataset or the model. In our study, we utilize ORDerly and PubChem as our primary databases. Although these are publicly accessible resources, their usage must strictly comply with established data protocols to prevent any potential misuse.
\bibliography{custom}

\appendix

\clearpage
\section{Appendix}
\label{sec:appendix}
\subsection{Training Details}
The CPT and SFT stages cost around 12 and 24 hours, respectively. During the RL phase, we deactivate the KL divergence and generate eight rollouts for group-based optimization. The clipping ratios are configured with dual thresholds of 0.28 and 0.2 for the upper and lower bounds, respectively. Due to the online filtering of DAPO, the whole RL stage costs around 72 hours for training.

\subsection{Evaluation Details}
\subsubsection{Evaluation Settings}
To ensure a fair comparison, we maintain a consistent prompting strategy across all baseline models for the results presented in Table \ref{tab:main_results}. Specifically, models are instructed to encapsulate the predicted molecular strings within \texttt{<SMILES>} and \texttt{</SMILES>} tags. Notably, for experiments involving Gemini-3-Flash \cite{googledeepmind2025gemini3flash}, we utilize the \texttt{gemini-3-flash-preview} version. To optimize efficiency and reduce computational resource consumption, we configure the model with a restricted thinking allowance, specifically setting the thinking budget to 128 tokens. This setting strike a balance between reasoning depth and inference overhead.

However, as distinct training paradigms and pre-training objectives often yield inconsistent output formats, we implement a robust heuristic parser to ensure the integrity of evaluation. This parser is engineered to identify and normalize various response patterns, including bounding box syntax \texttt{bbox\{\}}, tagged sequences \texttt{<SMILES>}, bolded text \texttt{**SMILES**}, and Markdown-style JSON blocks \texttt{\textasciigrave\textasciigrave\textasciigrave}. This standardization ensures robust extraction of model-generated outputs and facilitates a fair performance comparison across heterogeneous model architectures.

\subsubsection{Evaluation Metric}
The formula for Tanimoto similarity between two fingerprints (bit vectors) $A$ and $B$ is:
$$
T(A, B) = \frac{|A \cap B|}{|A \cup B|} = \frac{c}{a + b - c}
$$
where:
\begin{itemize}
    \item $a$ is the number of bits set in fingerprint $A$.
    \item $b$ is the number of bits set in fingerprint $B$.
    \item $c$ is the number of bits set in both $A$ and $B$ (the intersection).
\end{itemize}
We employ RDKit to compute molecular fingerprints for both the generated structures and the ground-truth molecules. Specifically, we report the Tanimoto Similarity, where Tani@1.0 denotes the percentage of samples achieving a similarity score of exactly 1.0. This metric serves as a proxy for exact chemical match, indicating that the generated SMILES string represents a structure identical to the ground truth.

\subsection{Dataset Analysis}
We analyze the statistics of our generated reasoning dataset in Table \ref{tab:reasoning_stats}. The results show that the average token counts for the three tasks range between 800 and 1000, which is significantly higher than those of existing direct-answering datasets such as ORDerly \cite{wigh2024orderly} and TinyChemVL \cite{zhao2025tinychemvl}.
\begin{table}[h]
  \centering
    \setlength{\tabcolsep}{3pt}
  \begin{tabular}{lccc}
    \toprule
    \textbf{Reasoning Data} & \textbf{Samples} & \textbf{Average} & \textbf{SD} \\
    \midrule
    Mol. Recognition & 95k & 811.10 & 307.44 \\
    Rxn. Recognition & 73k & 937.00 & 233.54 \\
    Rxn. Prediction & 192k & 1026.40 & 320.98 \\
    \bottomrule
  \end{tabular} 
  \vspace{-5pt}
  \caption{Statistics of reasoning thinking lengths in our constructed reasoning datasets. Token counts (mean and standard deviation) are calculated using the Qwen3-VL tokenizer.}
  \label{tab:reasoning_stats}
  \vspace{-15pt}
\end{table}

\subsection{Case Study}
The full responses and internal reasoning steps for the cases in Figure \ref{fig:showcase} are detailed in the \Cref{fig:mol_rec_internvl3_5_38b,fig:mol_rec_chemvlr_8b_sft,fig:mol_rec_chemvlr_8b,fig:rxn_pred_internvl3_5_38b,fig:rxn_pred_chemvlr_8b_sft_1,fig:rxn_pred_chemvlr_8b_sft_2,fig:rxn_pred_chemvlr_8b}.

\subsection{Prompts}
We provide the data generation prompts of reaction prediction tasks in Figure~\ref{fig:prompts_rxn_pred}.

\begin{figure*}[t]
\begin{tcolorbox}[colback=white, colframe=black, title=InternVL3.5-38B Molecular Recognition Response]
\small\texttt{The molecule shown in the image is a chemical compound with the following SMILES notation:\\\\SMILES: \\\textcolor{red}{CC(C)CN1C(=O)C2=C(C=CC=C2)C(C(=O)OCC(=O)N[C@@H]3CC[C@@H](C)C[C@H]3C)=N1}\\\\Let's break down the structure:\\\\1. Core Structure:\\   - The core structure is a bicyclic system consisting of a benzene ring fused to a pyrazolone ring.\\\\2. Substituents:\\   - The pyrazolone ring has a carbonyl group (C=O) at position 1.\\   - At position 3 of the pyrazolone ring, there is a side chain consisting of an ester linkage (OCC(=O)N) connected to an amide group.\\   - The amide group is attached to a cyclohexyl ring, which is further substituted with two methyl groups at positions 3 and 5.\\\\3. Stereochemistry:\\   - The cyclohexyl ring has specific stereochemistry indicated by the wedge and dash notation, showing the relative positions of the methyl groups.\\\\4. Functional Groups:\\   - Amide group (CONH)\\   - Ester group (COO)\\   - Carbonyl group (C=O)\\   - Alkyl groups (methyl and isopropyl)\\\\5. Detailed Breakdown:\\   - The SMILES notation starts with the isopropyl group (CC(C)C).\\   - The nitrogen (N) of the pyrazolone ring is connected to the isopropyl group.\\   - The carbonyl group (C=O) is attached to the nitrogen.\\   - The benzene ring is fused to the pyrazolone ring.\\   - The ester linkage (OCC(=O)N) connects the pyrazolone ring to the amide group.\\   - The amide group is attached to the cyclohexyl ring with specific stereochemistry.\\\\This compound is a synthetic derivative, often used in medicinal chemistry for its potential biological activities. The stereochemistry and functional groups suggest it may have specific interactions with biological targets.}
\end{tcolorbox}
\caption{Full response of InternVL3.5-38B on the Figure \ref{fig:showcase} molecular recognition tasks.}
\label{fig:mol_rec_internvl3_5_38b}
\end{figure*}

\begin{figure*}[t]
\begin{tcolorbox}[colback=white, colframe=black, title=ChemVLR-8B-SFT Molecular Recognition Response]
\small\texttt{<think>Deducing Molecular Structure:\\I'm analyzing the molecular image to identify its core components. I can clearly see a complex structure featuring a fused bicyclic system, a cyclohexane ring, and several functional groups. My initial focus is on breaking down this large molecule into recognizable substructures.\\\\Identifying the Fused Bicyclic System:\\I've identified a phthalazine derivative. This consists of a benzene ring fused to a six-membered heterocyclic ring containing two nitrogen atoms and a carbonyl group. Specifically, it appears to be a 1-oxo-1,2-dihydrophthalazine system. I'm noting the positions of the nitrogen atoms and the carbonyl group within this fused system.\\\\Identifying Substituents on the Phthalazine Core:\\On the phthalazine core, I observe a carboxylate group (-COO-) attached to one of the carbons of the heterocyclic ring. Additionally, one of the nitrogen atoms in the heterocyclic ring is substituted with an alkyl chain. This alkyl chain is a 2-methylpropyl group (isobutyl group).\\\\Analyzing the Cyclohexane Ring and its Substituents:\\I've identified a cyclohexane ring. This ring has two methyl groups attached to adjacent carbons. One of these carbons is also the point of attachment for a nitrogen atom, which is part of an amide linkage. I'm carefully noting the stereochemistry at these two substituted carbons on the cyclohexane ring, as indicated by the wedges and dashes in the image. It appears to be a trans-1,2-dimethylcyclohexyl group.\\\\Connecting the Fragments:\\The amide nitrogen from the cyclohexane derivative is linked to a carbonyl group, which is then connected to a methylene group (-CH2-). This methylene group is further linked to an oxygen atom, which in turn is part of the carboxylate group attached to the phthalazine core. This forms an ester linkage between the phthalazine carboxylic acid and the 2-oxoethyl group, which is then attached to the cyclohexyl amine.\\\\Constructing the SMILES String:\\ Now, I'm translating these structural observations into a SMILES string.\\1. Cyclohexane part: I'll start with the trans-1,2-dimethylcyclohexyl group. Based on the stereochemistry, it will be `C[C@@H]1CCC[C@H](C)C1`.\\2. Amide linkage: The nitrogen of the cyclohexyl group is attached to a carbonyl, then a methylene, then an oxygen. So, `NC(=O)COC(=O)`.\\3. Phthalazine part: The carboxylate is attached to the phthalazine. The phthalazine itself is a benzene ring fused to a pyrazine ring with a carbonyl. The nitrogen at position 2 is substituted with an isobutyl group. This translates to `c1nn(CC(C)C)c(=O)c2ccccc12`.\\4. Combining: I'm now assembling these parts, ensuring correct connectivity and ring closures. The `C(=O)` from the `COC(=O)` part connects to the `c1` of the phthalazine.\\\\Final Review:\\I'm performing a final check of the generated SMILES string against the mental image of the molecule. I'm verifying all atoms, bonds, stereochemistry, and ring closures. The string `C[C@@H]1CCC[C@H](NC(=O)COC(=O)c2nn(CC(C)C)c(=O)c3ccccc23)C1` accurately represents all the identified features.\\</think>\\\\<answer>\\<SMILES>\textcolor{red}{C[C@@H]1CCC[C@H](NC(=O)COC(=O)c2nn(CC(C)C)c(=O)c3ccccc23)C1}</SMILES>\\</answer>}
\end{tcolorbox}
\caption{Full response of ChemVLR-8B-SFT on the Figure \ref{fig:showcase} molecular recognition tasks.}
\label{fig:mol_rec_chemvlr_8b_sft}
\end{figure*}

\begin{figure*}[t]
\begin{tcolorbox}[colback=white, colframe=black, title=ChemVLR-8B Molecular Recognition Response]
\small\texttt{<think>\\Deducing Molecular Structure:\\I'm analyzing the molecular image to identify its core components. I can discern a complex structure featuring a cyclohexane ring, an amide linkage, an ester group, and a fused bicyclic system. My initial focus is on breaking down these major fragments.\\\\Analyzing the Cyclohexane Moiety:\\I've identified a six-membered saturated carbon ring, which is a cyclohexane. It appears to have two methyl substituents. One carbon atom on this ring is chiral and is directly attached to an amide nitrogen. I'm noting the stereochemistry at this carbon and the adjacent carbon bearing the second methyl group. This part of the molecule resembles a substituted 2,3-dimethylcyclohexyl group.\\\\Analyzing the Amide and Ester Linkages:\\Following the cyclohexane, I see an amide group (-NH-C(=O)-). The nitrogen of this amide is connected to the cyclohexane ring. The carbonyl carbon of the amide is linked to a methylene group (-CH2-), which in turn is connected to an oxygen atom. This oxygen atom forms an ester linkage with a carboxylic acid derivative. This suggests an acetate-like ester.\\\\Analyzing the Fused Bicyclic System:\\The ester group is attached to a complex bicyclic system. This system consists of a benzene ring fused to a six-membered heterocyclic ring. The heterocyclic ring contains two nitrogen atoms and a carbonyl group, indicating a phthalazine-1,4-dione derivative. One of the nitrogen atoms in this heterocyclic ring is substituted with an alkyl chain.\\\\Identifying the Alkyl Substituent:\\The alkyl chain attached to one of the nitrogen atoms in the phthalazine ring is a branched chain. It appears to be an isobutyl group, specifically a 2-methylpropyl group (-CH2-CH(CH3)2).\\\\Assembling the SMILES String:\\ Now, I'm integrating all these identified fragments into a single SMILES string. I'll start with the cyclohexane ring, incorporating its stereochemistry and methyl groups. Then, I'll attach the amide, followed by the ester linkage. Finally, I'll connect the ester to the phthalazine-1,4-dione core, ensuring the correct placement of the 2-methylpropyl substituent on the nitrogen. I'm carefully checking the ring closures and bond orders to ensure accuracy. The structure is consistent with a substituted phthalazine-1,4-dione ester of an N-(2,3-dimethylcyclohexyl)acetamide.\\</think> \\\\<answer>\\<SMILES>\textcolor{green}{C[C@@H]1CCC[C@H](NC(=O)COC(=O)c2nn(CC(C)C)c(=O)c3ccccc23)[C@@H]1C}</SMILES>\\</answer>}
\end{tcolorbox}
\caption{Full response of ChemVLR-8B on the Figure \ref{fig:showcase} molecular recognition tasks.}
\label{fig:mol_rec_chemvlr_8b}
\end{figure*}

\begin{figure*}[t]
\begin{tcolorbox}[colback=white, colframe=black, title=InternVL3.5-38B Reaction Prediction Response]
\small\texttt{Let's analyze the reaction step by step:\\\\ Step 1: Identify the Reactants\\- **Left Molecule:**  \\  The structure is a 2-chloro-4-(4-allyloxyphenyl)-2,5-dioxopyrrolidine (also known as a N-acyl-4-hydroxyphenyl glycine derivative with an allyl ether).\\- **Right Molecule:**  \\  The structure is a triethylamine derivative with two chloro groups (likely a reagent, but not directly reacting in a typical sense).\\\\ Step 2: Reaction Type\\- The presence of an allyl group (–CH2CH=CH2) attached to an oxygen (ether) and a base (triethylamine derivative) suggests a possible **base-induced elimination** or **nucleophilic substitution**.\\- However, in this context, the most likely reaction is **allyl ether cleavage** under basic conditions (a common reaction in peptide synthesis: the **Horner-Wadsworth-Emmons** or related elimination).\\\\ Step 3: Mechanism\\- The base (triethylamine, or a derivative) can deprotonate the $\alpha$-carbon to the carbonyl, leading to elimination of the allyl group as an alkene (forming a new double bond) and generating a phenolate intermediate.\\- Alternatively, the base can abstract a proton from the allyl group, leading to elimination and formation of a styryl (cinnamyl) structure.\\\\ Step 4: Product Structure\\- The allyl group (–O–CH2CH=CH2) is replaced by a hydrogen (due to elimination), resulting in a phenol (hydroxyl group directly on the aromatic ring).\\- The product is **2-chloro-4-(4-hydroxyphenyl)-2,5-dioxopyrrolidine**.\\\\ Step 5: Write the SMILES\\- The product is:\\  - A pyrrolidine-2,5-dione (succinimide) with a chlorine at the 2-position and a 4-hydroxyphenyl group at the 4-position.\\\\ SMILES of Product\\- The core is: O=C1CC(Cl)C(=O)N1\\- The 4-position is attached to a phenyl ring with a hydroxyl at the para position: c1ccc(O)cc1\\- So, the full SMILES is:  \\  <SMILES>\textcolor{red}{O=C1CC(Cl)C(=O)N1C1=CC=C(O)C=C1}</SMILES>}
\end{tcolorbox}
\caption{Full response of InternVL3.5-38B on the Figure \ref{fig:showcase} reaction prediction tasks.}
\label{fig:rxn_pred_internvl3_5_38b}
\end{figure*}

\begin{figure*}[t]
\begin{tcolorbox}[colback=white, colframe=black, title=ChemVLR-8B-SFT Reaction Prediction Response]
\small\texttt{<think>\\Starting the Analysis\\I will begin by analyzing the provided image, focusing on identifying the reactants, reagents, and solvent. My first step is to discern the functional groups present in each molecule to understand their potential reactivity.\\\\Examining the Reactants\\I've identified the main reactant as a complex molecule featuring an allyloxy group (`C=CCO-`) attached to a phenyl ring, which in turn is substituted with a succinimide derivative. The succinimide ring contains a chlorine atom. Specifically, the structure is `C=CCOc1ccc(N2C(=O)CC(Cl)C2=O)cc1`. The key reactive sites I'm considering are the alkene (allyl group) and the chlorine atom on the succinimide ring.\\\\Analyzing Reagents and Solvent\\The reagents are triethylamine (`CCN(CC)CC`), a common non-nucleophilic base, and the solvent is dichloromethane (`ClCCl`), a typical aprotic solvent. The presence of a base suggests that an acidic proton might be removed, or an elimination reaction could occur.\\\\Hypothesizing the Mechanism\\Given the structure of the reactant, particularly the allyloxy group and the chlorine on the succinimide, I'm considering an intramolecular cyclization. The product structure, which I'm aiming to derive, often involves the formation of a new ring. The presence of a base (triethylamine) strongly points towards an elimination reaction, likely an E2, to generate a reactive intermediate.\\\\Mapping the Transformation\\My working hypothesis is that the triethylamine will deprotonate an acidic proton on the succinimide ring, specifically the proton alpha to the chlorine atom. This deprotonation will lead to an E2 elimination of HCl, generating a vinyl chloride intermediate. This vinyl chloride, being an alkene, will then undergo an intramolecular cyclization with the allyl group. The double bond of the vinyl chloride will react with the double bond of the allyl group, forming a new six-membered ring. This type of reaction is a known method for synthesizing dihydrobenzofuran derivatives.\\\\Refining Reaction Prediction\\I'm now focusing on the specific cyclization. The vinyl chloride intermediate, formed by the elimination of HCl from the succinimide, will have a double bond. The allyl group, `C=CCO-`, will act as a nucleophile. The terminal carbon of the allyl group will attack the carbon of the vinyl chloride that is directly attached to the succinimide ring. This will form a new C-C bond, and the oxygen of the allyl group will then cyclize onto the phenyl ring, forming a dihydrobenzofuran ring system. The succinimide moiety will remain attached to the carbon that was originally part of the succinimide ring and is now part of the newly formed dihydrobenzofuran ring.\\\\Finalizing Product Structure\\Based on the refined mechanism, the product will be a dihydrobenzofuran derivative. The phenyl ring from the starting material will be fused to a five-membered dihydrofuran ring. The oxygen of this dihydrofuran ring will be connected to the phenyl ring and to a carbon that was originally part of the allyl group. This carbon will also be connected to another carbon, which in turn is connected to the carbon that was part of the succinimide ring. The succinimide group will remain attached to this latter carbon. I will now construct the SMILES string for this predicted structure.\\</think>}
\end{tcolorbox}
\caption{Full response of ChemVLR-8B on the Figure \ref{fig:showcase} molecular recognition tasks.}
\label{fig:rxn_pred_chemvlr_8b_sft_1}
\end{figure*}

\begin{figure*}[t]
\begin{tcolorbox}[colback=white, colframe=black, title=ChemVLR-8B-SFT Reaction Prediction Response]
\small\texttt{
<answer>\\Thought Process Analysis\\1.  **Analyze Reactants:**\\    *   **Main Reactant:** `C=CCOc1ccc(N2C(=O)CC(Cl)C2=O)cc1`\\        *   This molecule contains an allyloxy group (`C=CCO-`) attached to a phenyl ring.\\        *   The phenyl ring is also substituted with a succinimide derivative (`N2C(=O)CC(Cl)C2=O`).\\        *   The succinimide ring has a chlorine atom on one of its carbons, specifically at the 3-position (alpha to the nitrogen and beta to the carbonyls).\\        *   Key reactive sites: The alkene (allyl group) and the chlorine atom on the succinimide ring.\\\\2.  **Analyze Reagents and Solvent:**\\    *   **Reagent:** `CCN(CC)CC` (Triethylamine, Et3N). This is a strong, non-nucleophilic base. Its primary role is typically to deprotonate acidic protons or to facilitate elimination reactions.\\    *   **Solvent:** `ClCCl` (Dichloromethane, DCM). A common aprotic solvent, generally inert to the reaction.\\\\3.  **Predict Reaction Mechanism:**\\    *   The presence of a base (Et3N) and a chlorine atom on the succinimide ring suggests an elimination reaction.\\    *   The succinimide ring has a chlorine at the 3-position. The proton at the 4-position (alpha to the chlorine and beta to the carbonyls) is acidic due to the electron-withdrawing effects of the adjacent carbonyls and the chlorine.\\    *   **Step 1: E2 Elimination.** Triethylamine will deprotonate the acidic proton at the 4-position of the succinimide ring. This will lead to an E2 elimination of HCl, forming a vinyl chloride intermediate. The double bond will be between the 3- and 4-positions of the succinimide ring.\\        *   Intermediate: `C=CCOc1ccc(N2C(=O)C(Cl)=C2)cc1` (where the double bond is between C3 and C4 of the succinimide).\\    *   **Step 2: Intramolecular Cyclization.** The vinyl chloride intermediate now has a double bond. The allyl group (`C=CCO-`) is positioned to undergo an intramolecular cyclization.\\        *   The terminal carbon of the allyl group (the one furthest from the oxygen) will attack the carbon of the vinyl chloride that is directly attached to the succinimide ring (C3).\\        *   This forms a new C-C bond, creating a six-membered ring.\\        *   The oxygen of the allyl group then cyclizes onto the phenyl ring, forming a five-membered dihydrobenzofuran ring.\\        *   This is a type of intramolecular hydroalkoxylation or a formal [4+2] cycloaddition, leading to the formation of a dihydrobenzofuran derivative.\\\\4.  **Determine Product Structure:**\\    *   The phenyl ring from the starting material remains intact.\\    *   A new five-membered dihydrofuran ring is formed, fused to the phenyl ring. The oxygen of this ring comes from the allyloxy group.\\    *   The succinimide moiety remains attached to the carbon that was originally part of the succinimide ring and is now part of the newly formed dihydrobenzofuran ring.\\    *   The resulting structure is a 2,3-dihydrobenzofuran derivative with the succinimide group attached to the 2-position of the dihydrofuran ring.\\\\5.  **Generate SMILES String:**\\    *   Based on the predicted structure, the SMILES string for the product is: `O=C1CC(c2cc3c(o2)CCCO3)C(=O)N1c1ccc(O)cc1`.\\    *   Let's break down the product SMILES to confirm it matches the derived structure:\\        *   `O=C1CC(...)C(=O)N1`: This represents the succinimide ring.\\        *   `c2cc3c(o2)CCCO3`: This represents the 2,3-dihydrobenzofuran ring system. `o2` is the oxygen of the dihydrofuran, `c2` is the carbon of the dihydrofuran fused to the phenyl ring, and `CCCO3` forms the rest of the dihydrofuran ring.\\        *   The succinimide is attached to the carbon `C` in `CC(...)`, which is the 2-position of the dihydrobenzofuran.\\        *   `c1ccc(O)cc1`: This is the phenyl ring, which is part of the dihydrobenzofuran system.\\\\The final product is 2-(2,3-dihydrobenzofuran-2-yl)-1H-isoindole-1,3(2H)-dione.\\\\The final answer is:\\<SMILES>\textcolor{red}{O=C1CC(c2cc3c(o2)CCCO3)C(=O)N1c1ccc(O)cc1}</SMILES>\\</answer>}
\end{tcolorbox}
\caption{Full response of ChemVLR-8B-SFT on the Figure \ref{fig:showcase} reaction prediction tasks.}
\label{fig:rxn_pred_chemvlr_8b_sft_2}
\end{figure*}

\begin{figure*}[t]
\begin{tcolorbox}[colback=white, colframe=black, title=ChemVLR-8B Reaction Prediction Response]
\small\texttt{<think>\\Starting the Analysis\\I've started dissecting the provided chemical information, which I'm treating as an image. My first step is to identify the main reactant, any reagents, and the solvent present. This will allow me to understand the chemical environment and potential reaction pathways.\\\\Examining the Reactants and Reagents\\According to the given chemical information, I see one primary organic molecule and one organic base.\\The main organic molecule is `C=CCOc1ccc(N2C(=O)CC(Cl)C2=O)cc1`. This molecule contains an allyloxy group, a phenyl ring, and a succinimide ring. Crucially, the succinimide ring has a chlorine atom attached to one of its carbons.\\The second component is `CCN(CC)CC`, which is triethylamine. This is a common non-nucleophilic organic base.\\The solvent is `ClCCl`, dichloromethane, a common aprotic solvent.\\\\Hypothesizing the Mechanism\\My immediate focus is on the chlorine atom attached to the succinimide ring. Chlorine atoms on carbon, especially in cyclic systems or adjacent to carbonyls, can be susceptible to elimination reactions. Given the presence of a strong organic base (triethylamine), an elimination reaction, specifically a dehydrohalogenation, is highly probable. The base will abstract a proton, leading to the expulsion of the chloride ion and the formation of a carbon-carbon double bond.\\\\Mapping the Transformation\\I'm now mapping out the likely transformation. The triethylamine will act as a base, abstracting a proton from the carbon adjacent to the chlorine atom on the succinimide ring. This will lead to the formation of a double bond between that carbon and the carbon that originally held the chlorine. The chloride ion will depart as a leaving group. This process will convert the saturated succinimide ring into an unsaturated one, specifically a maleimide derivative. The rest of the molecule, including the allyloxy group and the phenyl ring, should remain unchanged under these conditions.\\\\Analyzing the Transformation\\I've completed the analysis of the reaction. The key transformation is the dehydrohalogenation of the chlorinated succinimide. The triethylamine will deprotonate an alpha-hydrogen, leading to the elimination of HCl and the formation of a carbon-carbon double bond within the succinimide ring. This will result in a maleimide structure. The allyloxy-substituted phenyl group will remain intact.\\\\Finalizing Product Structure\\I'm now confident about the structure of the product. The chlorine atom on the succinimide ring will be replaced by a double bond, forming a maleimide derivative. The rest of the molecule, including the allyloxy group and the phenyl ring, will be preserved. I will now generate the SMILES string for this predicted product.\\</think>\\\\\\\\<answer>\\Predicted Product SMILES:\\<SMILES>\textcolor{green}{C=CCOc1ccc(N2C(=O)C=CC2=O)cc1}</SMILES>\\</answer>}
\end{tcolorbox}
\caption{Full response of ChemVLR-8B on the Figure \ref{fig:showcase} reaction prediction tasks.}
\label{fig:rxn_pred_chemvlr_8b}
\end{figure*}

\begin{figure*}[ht]
\begin{tcolorbox}[colback=white, colframe=black, title=Prompts for Generating the Reasoning Process of Reaction Prediction tasks]
\small\texttt{As a chemistry expert, you are given a reaction image, please analyze the provided SMILES. You need to give the thinking process and final answer in the <think> and </think>, <answer> and </answer> process. \\
Here is an example, the instruction of the example is \\
Assume you are a chemistry expert. Based on the reactants, reagents, and solvents shown in the image, predict the SMILES of the product. To prevent you from generating incorrect answers, I will provide the product of the reaction. Assume you do not know the product and use your thinking and imagination to derive the SMILES of the product. \\
\{\textcolor{blue}{Expert Demos}\} \\
Now, give you the SMILES of reactant, agent and solvent to prevent recognition error. You should assume that you are really analyze from the image. SMILES and corresponding IUPAC names are as follows:\\
Reactant:\{"SMILES": \textcolor{blue}{SMILES}, "IUPAC": \textcolor{blue}{IUPAC}, "Function Groups": \textcolor{blue}{Function Groups}\}.\{"SMILES": \textcolor{blue}{SMILES}, "IUPAC": \textcolor{blue}{IUPAC}, "Function Groups": \textcolor{blue}{Function Groups}\}
\\
Solvents/Agents:\{"SMILES": \textcolor{blue}{SMILES}, "IUPAC": \textcolor{blue}{IUPAC}, "Function Groups": \textcolor{blue}{Function Groups}\}.\{"SMILES": \textcolor{blue}{SMILES}, "IUPAC": \textcolor{blue}{IUPAC}, "Function Groups": \textcolor{blue}{Function Groups}\} \\
To prevent you from generating incorrect answers, I will provide the product of the reaction. Assume you do not know the product and use your thinking and imagination to derive the SMILES of the product. The product of the reaction shown is \{"SMILES": \textcolor{blue}{SMILES}, "IUPAC": \textcolor{blue}{IUPAC}, "Function Groups": \textcolor{blue}{Function Groups}\} \\
Please follow the example to give the detailed process and final answer. Your goal is to generate the thought process you would have while observing the chemical reactant, solvent and agent image. Thus, assuming you are given the image only. Do not presented the SMILES string at the beginning of thinking, direct analyzing the image at the beginning and obtain the products finally. Such as 'I will (begin with/start by) analyzed/dissected the provided image' or 'According to the given chemical image'. Do not think too long.
}
\end{tcolorbox}
\caption{The prompt to generate the reasoning process of reaction prediction.}
\label{fig:prompts_rxn_pred}
\end{figure*}

\end{document}